\documentclass[runningheads]{llncs}
\usepackage{graphicx}
\usepackage{amsmath}
\usepackage{wrapfig}

\begin{document}
\title{Searching for Efficient Architecture for Instrument Segmentation in Robotic Surgery}

\author{Daniil Pakhomov$^{1}$ and Nassir Navab$^{1,2}$}
\institute{$^{1}$ Johns Hopkins University, USA \\
$^{2}$ Technische Universität München, Germany}

\titlerunning{ }

\authorrunning{D. Pakhomov et al.}
%
%
%
%
%
\maketitle              
\begin{abstract}

Segmentation of surgical instruments is an important problem in robot-assisted surgery: it is a crucial step towards full instrument pose estimation and is directly used for masking of augmented reality overlays during surgical procedures. Most applications rely on accurate real-time segmentation of high-resolution surgical images. While previous research focused primarily on methods that deliver high accuracy segmentation masks, majority of them can not be used for real-time applications due to their computational cost. In this work, 
we design a light-weight and highly-efficient deep residual architecture which is tuned to perform real-time inference of high-resolution images. To account for  reduced accuracy of the discovered light-weight deep residual network and avoid adding any additional computational burden, we perform a differentiable search over dilation rates for residual units of our network. We test our discovered architecture on the EndoVis 2017 Robotic Instruments dataset and verify that our model is the state-of-the-art in terms of speed and accuracy tradeoff with a speed
of up to 125 FPS on high resolution images.
\end{abstract}
\section{Introduction}

\begin{figure*}
\centering
		\includegraphics[width=1\textwidth]{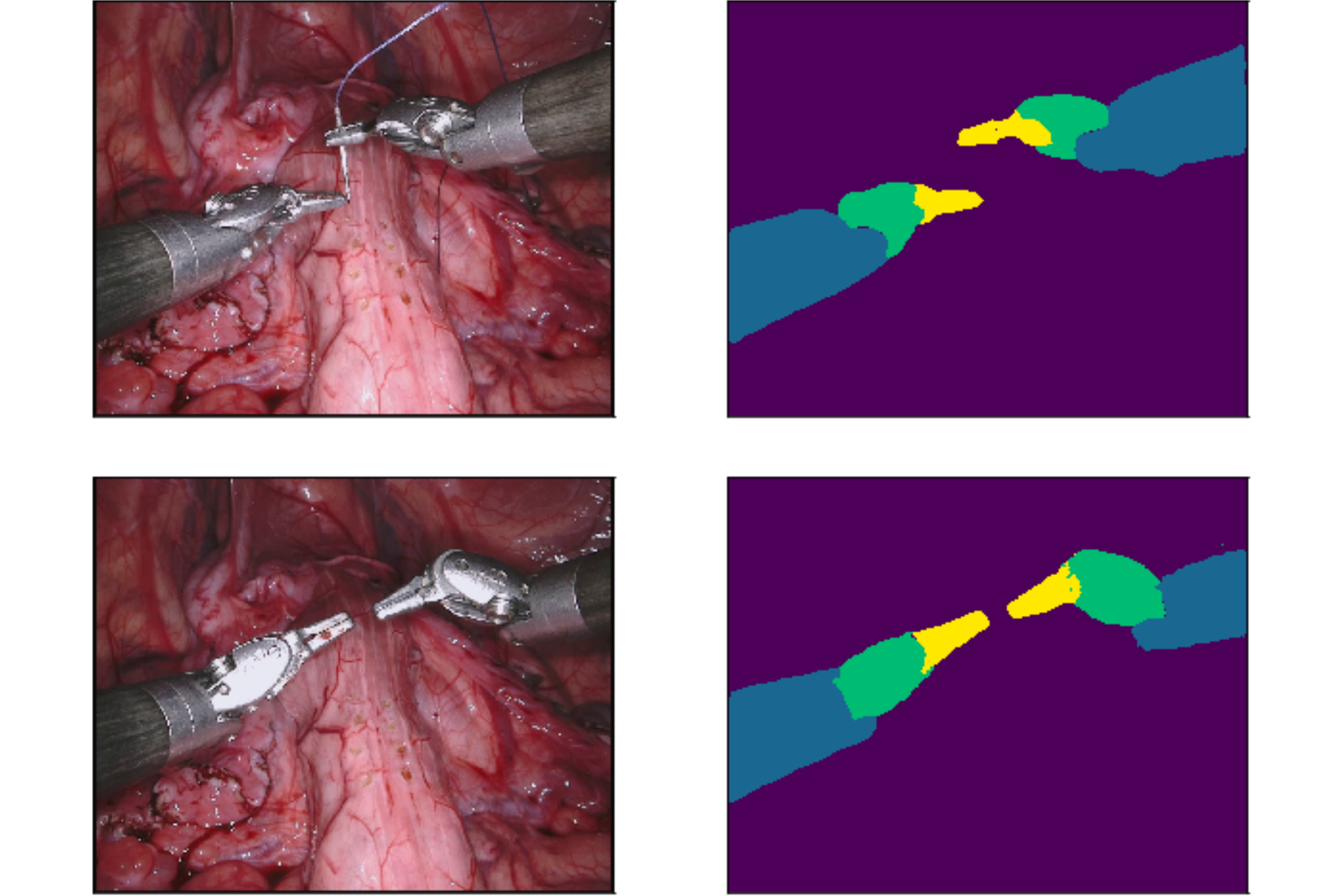}
\caption{\label{fig:demo} Multi-class instrument segmentation results delivered by our method on sequences from validational subset of Endovis 2017 dataset.}
\end{figure*}

Robot-assisted Minimally Invasive Surgery (RMIS) provides a surgeon with improved control, facilitating procedures in confined and difficult to access anatomical regions. However, complications due to the reduced field-of-view
provided by the surgical camera limit the surgeon’s ability to self-localize. Computer assisted interventions (CAI) can help a surgeon by integrating additional information. For example, overlaying pre- and intra-operative imaging with the surgical console can provide a surgeon with valuable information which can improve decision making during complex procedures~\cite{taylor2008medical}. Integrating this data is a complex
task and involves understanding relations between the patient anatomy, operating instruments and surgical camera. Segmentation of the instruments in the camera images is a crucial component of this process and can be used to prevent rendered overlays from occluding the instruments while providing crucial input to instrument tracking frameworks~\cite{pezzementi2009articulated,allan2014d}.

There has been a significant development in the field of instrument segmentation~\cite{pakhomov2019deep,shvets2018automatic,laina2017concurrent} based on recent advancement in deep learning~\cite{he2016deep,chen2016deeplab}. Additionally, release of new datasets for surgical tools segmentation with high-resolution challenging images further improved the field by allowing methods to be more rigorously tested~\cite{allan20192017}. While the state-of-the-art methods show impressive pixel accurate results~\cite{pakhomov2019deep,shvets2018automatic}, the inference time of these methods makes them unsuitable for real-time applications. In our work we address this problem and present a method that has a faster than real time inference time while delivering high quality segmentation masks.

\section{Method}

In this work, we focus on the problem of surgical instrument segmentation. Given an input image, every pixel has to be classified into one of $C$ mutually exclusive classes. We consider two separate tasks with increasing difficulty: binary tool segmentation and multi-class tool segmentation. In the first case, each pixel has to be classified into $C=2$ classes as belonging to surgical background or instrument. In the second task, into $C=4$ classes, namely tool's shaft, wrist, jaws and surgical background. Our method is designed to perform these tasks efficiently while delivering high quality results. First, we discuss the previous state-of-the-art method based on dilated residual networks (Section \ref{sec:dilated}) and highlight the major factor that makes it computationally expensive. Then, we present light residual networks (Section \ref{sec:light}) that allow to solve this problem and make the method faster. To account for reduced accuracy we introduce a search for optimal dilation rates in our model (Section \ref{sec:dilation_search}) which allows to improve its accuracy.

\subsection{Dilated Residual Networks}
\label{sec:dilated}
We improve upon previous state-of-the-art approach~\cite{pakhomov2019deep} based on dilated residual network that employs ResNet-18 (see Fig.~\ref{fig:method}), a deep residual network pretrained on ImageNet dataset. The network is composed of four successive residual blocks, each one consisting of two residual units of "basic" type~\cite{he2016deep}. The average pooling layer is removed and stride is set to one in the last two residual blocks responsible for downsampling, subsequent convolutional layers are dilated with an appropriate rate as suggested in~\cite{pakhomov2019deep,chen2016deeplab}. This allows to obtain predictions that are downsampled only by a factor of 8$\times$ (in comparison to the original downsampling of 32$\times$) which makes the network work on a higher resolution features maps (see Fig.~\ref{fig:method}) and deliver finer predictions~\cite{chen2016deeplab,pakhomov2019deep}.

The aforementioned method delivers very accurate segmentation masks but is too computationally expensive for real-time applications (see. Table~\ref{table:quantative_results}). The main reason for this is that it uses deep residual classification models pretrained on ImageNet that usually have great number of filters at the last layers: when the model is being transformated into dilated residual network, the downsampling operations at the last two residual blocks are removed forcing the convolutional layers to operate on the input that is spatially bigger by a factor of two and four respectively~\cite{chen2016deeplab,pakhomov2019deep} (see Fig.~\ref{fig:method}). Even the most shallow deep residual network ResNet-18 is too computationally expensive when converted into dilated residual network.
This motivates us to create a smaller deep residual network that can be pretrained on the imageNet and converted into dilated residual model that exhibits improved running time.

\subsection{Light Residual Networks}
\label{sec:light}
\begin{figure*}
\centering
		\includegraphics[width=1\textwidth]{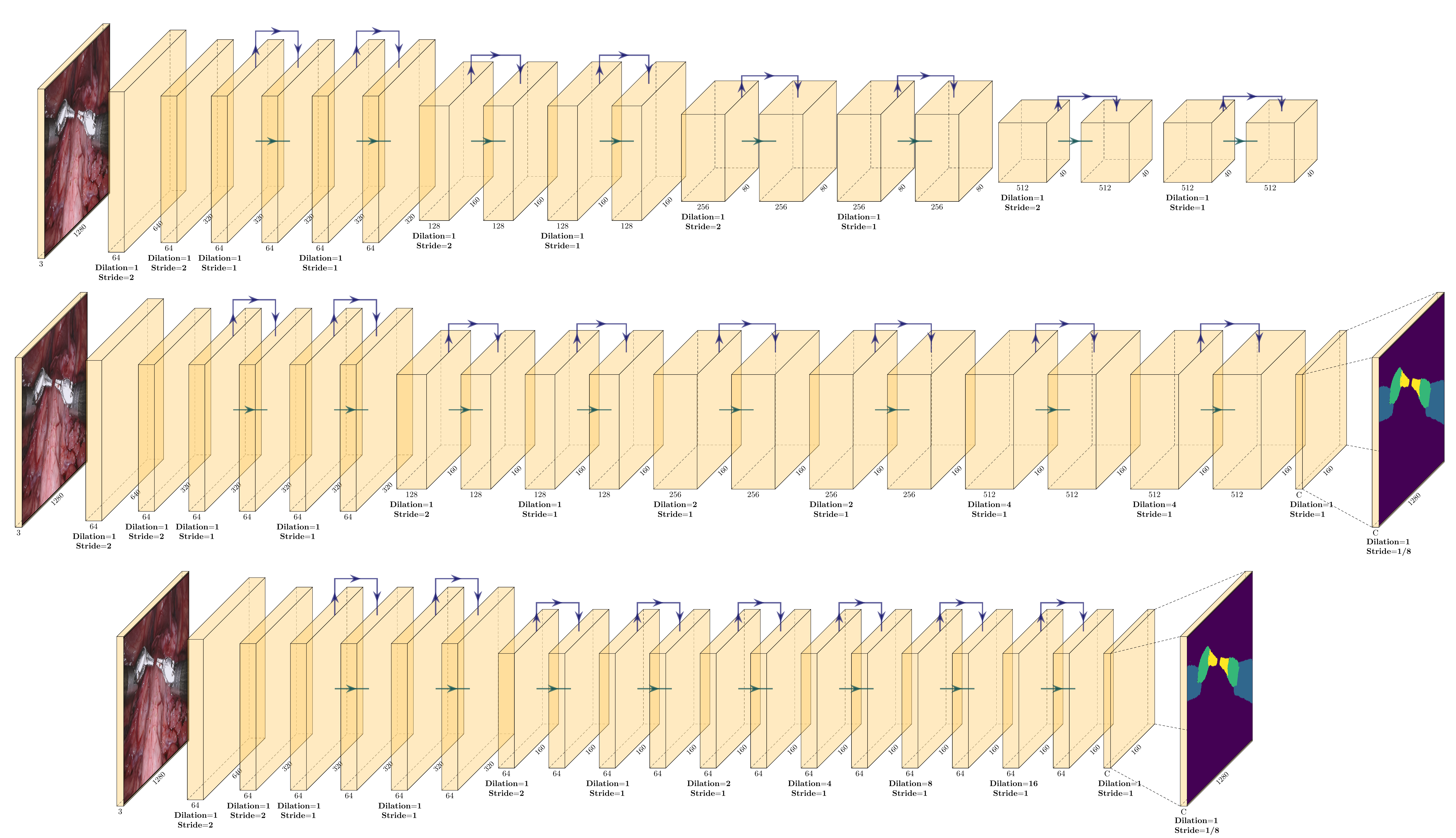}
\caption{\label{fig:method} \textbf{Top row:} simplified ResNet-18 before being converted into dilated fully convolutional network (FCN). \textbf{Middle row:} ResNet-18 after being converted into FCN following the method of~\cite{pakhomov2019deep}. Since the stride was set to one in two layers and the number of channels was left the same, convolutional layers are forced to work on a greater resolution, therefore, computational overhead is substantially increased, making the network unable to deliver real-time performance. \textbf{Bottom row:} Our light-weight ResNet-18 being converted into dilated FCN and dilation rates are set to the values found during our differentiable search. The decreased number of channels makes the network fast while the found dilation rates increase its accuracy without any additional parameters or computational overhead. Green arrows represent the residual unit of a "basic" type, black arrows represent skip connections. First two layers does not have any residual units and are simplified in the figure. Dashed lines represent the bilinear upsampling operation. The figure is better viewed in a pdf viewer in color and zoomed in for details. The figure in bigger resolution is available in the supplementary material.}
\end{figure*}

In order to solve the aforementioned problem, we introduce a light residual network, a deep residual model which satisfies the requirements:

\begin{itemize}
    \item \textbf{Low latency:} exhibits low latency inference on high-resolution images when converted into dilated residual network. This is achieved by reducing the number of filters in the last stages of the network (see Fig.~\ref{fig:method}). In the original resnet networks~\cite{he2016deep}, the number of filters in the last stages is considerable: after being converted into dilated versions, these stages experience increase in the computational price by a factor of two and four~\cite{chen2016deeplab}. Since last stages are the biggest factor responsible for the increased inference time, we decrease the number of channels in them. This significantly decreases the inference time (see Table~\ref{table:quantative_results}). 
    While we noticed a considerable decrease in the accuracy of the model on the ImageNet dataset, the performance on the segmentation dataset only moderately decreased. We attribute it to the fact the number of filters in the last layers is of significant importance for the imagenet classification task because it needs to differentiate between one thousand classes compared to only four in our case.
    \item \textbf{Low GPU memory consumption:} the memory requirements of the network allow it to be trained with optimal batch and image crop sizes on single GPU device. Dilated residual networks consume a considerable amount of memory when trained for segmentation task~\cite{chen2016deeplab}, while still requiring relatively big batch size and crop size~\cite{cheng2019panoptic}. Similar to the previous section, the biggest factor responsible for the memory comsumption again involves the last stages of the network~\cite{chen2016deeplab}. Since every activation has to be stored in memory during backpropagation~\cite{chen2016training}, the layers that generate the biggest activations are contributing the most to the increased memory consumption. The last layers of the residual network work on increased spatial resolution after being converted to the dilated residual network and have more channels than other layers. By decreasing the number of channels in the last layers, we solve the problem and are able to train the model with sufficient batch size and image crop size~\cite{cheng2019panoptic}.
    \item \textbf{ImageNet pretraining:} the network is pretrained on ImageNet dataset which was shown to be essential for good performance on small segmentation datasets like Endovis~\cite{shvets2018automatic}. We pretrain all our models on the ImageNet dataset following the parameters suggested in~\cite{he2016deep}.
\end{itemize}

We present two versions of Light ResNet-18 named Light ResNet-18-v1 and Light ResNet-18-v2 with number of channels in the last layers set to $64$ and $32$ respectively. Second version exhibits improved runtime speed at the expense of decreased accuracy on the segmentation task. All models were pretrained on imageNet dataset and then converted into dilated residual networks following~\cite{pakhomov2019deep}. After being converted, the network were trained on the Endovis 2017~\cite{allan20192017} segmentation dataset for binary and multi-class instrument segmentation tasks. While being fast, the networks exhibit reduced accuracy compared to state-of-the-art methods (see. Table~\ref{table:quantative_results}). In order to account for that and improve the accuracy without reducing the speed of the network, we search for optimal dilation rates for each layer.

\subsection{Searching for Optimal Dilation Rates}
\label{sec:dilation_search}

\begin{figure*}
\centering
		\includegraphics[width=0.5\textwidth]{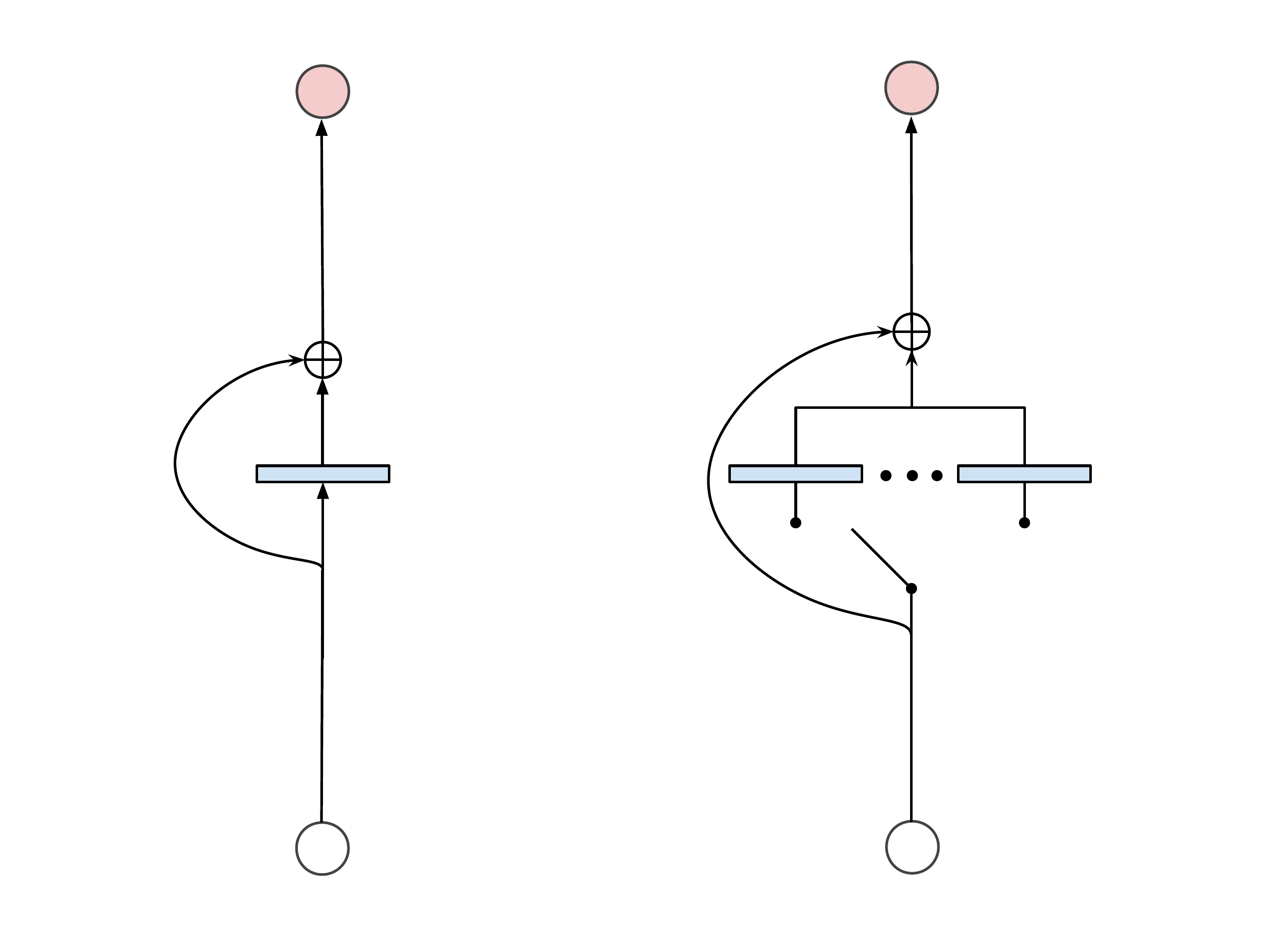}
\caption{\label{fig:residual_unit} \textbf{Left:} Simplified residual unit, a building block of deep residual networks which are usually formed as a sequence of residual units. \textbf{Right:} A group of residual units with different dilation rates combined with a discrete decision gate which forces the network to choose only one of residual units depending on which of them leads to a better overall performance of the network. The gate is controlled by a variable which receives gradients during training. During optimization each layer has its own set of residual units and gate variables and by the end of training only the residual units with dilation rates that perform best are left (other choices can be safely removed). We perform search for dilation rates only for the last four residual units of our network.}
\end{figure*}

In order to improve accuracy of our model further but avoid adding any new parameters or additional computational overhead~\cite{he2017learning}, we search for optimal integer dilation rates for each residual unit. Since trying out all possible combinations of dilation rates and retraining a model each time is infeasible, we formulate the problem of dilation rate search as an optimization problem~\cite{xie2018snas}.

First, we update residual units of our model. Original residual unit~\cite{he2016deep} can be expressed in a form (see Figure ~\ref{fig:residual_unit}):

\begin{equation}
x_{l+1} = x_{l} + \mathcal{F}(x_{l})
\end{equation}

where $x_{l}$ and $x_{l+1}$ are input and output of the $l$-th unit, and $\mathcal{F}$ is a residual function.

In order to allow our network to choose residual units with different dilation rates, we introduce gated residual unit which has $N$ different residual connections with different dilation rates:

\begin{gather}
x_{l+1} = x_{l} + \sum_{i=0}^{N} \textbf{Z}_{i} \cdot \mathcal{F}_i(x_{l}) \\
\sum_{i=0}^{N} \textbf{Z}_i = 1 \\
\forall i~\textbf{Z}_i \in \{0, 1\}
\end{gather}

where $\mathbf{Z}$ is a gate that decides which residual connection to choose and is represented with discrete one-hot-encoded categorical variable, $N$ is equal to the number of dilation rates that we consider. In order to be able to search for the best dilation rates, we need to also optimize the gate variable $\mathbf{Z}$ which is not differentiable since it represents a hard decision. But it is possible to obtain estimated gradients of such hard decisions by introducing petrtubations in the system in the form of noise~\cite{xie2018snas,maddison2016concrete,jang2016categorical,veit2017convolutional}. To be more specific, we use recent work that introduces differentiable approximate sampling mechanism for categorical variables based on a Gumbel-Softmax distribution (also known as a concrete distribution)~\cite{jang2016categorical,maddison2016concrete,veit2017convolutional}. We use Gumbel-Softmax to relax the discrete distribution to be continuous and differentiable with reparameterization trick:
\begin{equation}
\bar{\mathbf{Z}}_i = softmax( (\log\alpha_i + G_i) / \tau )
\label{eq:GumbelSoftmax}
\end{equation}

Where $G_i$ is an $i$th Gumbel random variable, $\bar{\mathbf{Z}}$ is the softened one-hot random variable which we use in place of $\mathbf{Z}$, $\alpha_i$ is a parameter that controls which residual unit to select and is optimized during training, $\tau$ is the temperature of the softmax, which is steadily annealed to be close to zero during training.

We update last four layers or our network to have gated residual units. Each of them allows the network to choose from a predefined set of dilation rates which we set to $\{1, 2, 4, 8, 16\}$. We train the network on the Endovis 2017 dataset by optimizing all weights including the $\alpha_i$ variables which control the selection of dilation rates. Upon convergence, the best dilation rates are decoded from the $\alpha_i$ variables. Discovered dilation rates can be seen at the Fig.~\ref{fig:method}. Next we train the residual network with specified dilation rates and original residual units.

\subsection{Training}

After the optimal dilation rates are discovered we update Light ResNet-18-v1 and ResNet-18-v2 to use them. Light ResNet-18-v1 and ResNet-18-v2 were first pretrained on Imagenet. We train networks on the Endovis 2017 train dataset~\cite{allan20192017}. During training we recompute the batch normalization statistics~\cite{ioffe2015batch,cheng2019panoptic}. We optimize normalized pixel-wise cross-entropy loss~\cite{chen2016deeplab} using Adam optimization algorithm. Random patches are cropped from the images~\cite{chen2016deeplab} for additional regularization. We employ crop size of $799$. We use the ‘poly’ learning rate policy with an initial learning rate of $0.001$~\cite{cheng2019panoptic}. The batch size is set to $32$.

\section{Experiments and Results}

We test our method on the EndoVis 2017 Robotic Instruments dataset~\cite{allan20192017}. There are $10$ $75$-frame sequences in the test dataset that features 7 different robotic surgical instruments~\cite{allan20192017}. 
Samples from the dataset and qualitative results of our method are depicted in Fig.~\ref{fig:demo}. We report quantitative results in terms of accuracy and inference time of our method in the Table.~\ref{table:quantative_results} and Table.~\ref{table:quantative_results_2}.

\begin{table}
\caption{Quantitative results of our method compared to other approaches in terms of accuracy (measured using mean intersection over union metric) and latency (measured in miliseconds). Latency was measured for an image of input size $1024\times1280$ using NVIDIA GTX 1080Ti GPU. It can be seen that our light backbone allows for significantly decreased inference time, while learnt dilations help to improve decreased accuracy.}
\label{table:quantative_results}
\centering   
\begin{tabular}{|c | c | c | c | c | c | c | c | c | c|}
\hline\noalign{\smallskip}
&
\multicolumn{2}{c}{Binary segmentation} &
\multicolumn{2}{|c|}{Parts segmentation} \\
\hline
Model & IOU & Time & IOU & Time\\
\hline
TernausNet-16~\cite{shvets2018automatic,allan20192017} & 0.888 & 184 ms & 0.737 & 202 ms\\
Dilated ResNet-18~\cite{pakhomov2019deep} & \textbf{0.896} & 126 ms & \textbf{0.764} & 126 ms\\
Dilated Light ResNet-18-v1 & 0.821 & 17.4 ms  & 0.728 & 17.4 ms\\
Light ResNet-18-v1 w/ Learnt Dilations & 0.869 & 17.4 ms & 0.742 & 17.4 ms \\
Dilated Light ResNet-18-v2 & 0.805 & \textbf{11.8 ms} & 0.706 & \textbf{11.8 ms }\\
Light ResNet-18-v2  w/ Learnt Dilations & 0.852 & \textbf{11.8 ms} & 0.729 & \textbf{11.8 ms}\\
\hline
\end{tabular}
\end{table}

\begin{table}
\caption{Latency of our method as measured on a modern NVIDIA Tesla P100 GPU for an image of input size $1024\times1280$. We can see that fastest of our models is able to work at $125$ frames per second.}
\label{table:quantative_results_2}
\centering   
\begin{tabular}{|c | c | c | c | c | c | c | c | c | c|}
\hline\noalign{\smallskip}
&
\multicolumn{2}{c}{Binary segmentation} &
\multicolumn{2}{|c|}{Parts segmentation} \\
\hline
Model & IOU & Time & IOU & Time\\
\hline
Light ResNet-18-v1  w/ Learnt Dilations & 0.869 & 11.5 ms & 0.742 & 11.5 ms \\
Light ResNet-18-v2  w/ Learnt Dilations & 0.852 & \textbf{7.95 ms} & 0.729 & \textbf{7.95 ms}\\
\hline
\end{tabular}
\end{table}

As it can be seen our method is able to deliver pixel accurate segmentation while working at an extremely fast frame rate of up to 125 FPS.

\section{Discussion and Conclusion}

In this work, we propose a method to perform real-time robotic tool segmentation on high resolution images. This is an important task,
as it allows the segmentation results to be used for applications that require low latency, for example, preventing rendered overlays from occluding the instruments or estimating the pose of a tool~\cite{allan2014d}. We introduce a lightweight deep residual network to model the mapping from the raw images to the segmentation maps that is able to work at high frame rate. Additionally, we introduce a method to search for optimal dilation rates for our lightweight model, which improves its accuracy in binary tool and instrument part segmentation. Our results show the benefit of our method for this task and also provide a solid baseline for the future work.

\end{document}